\documentclass{llncs}

\usepackage{graphicx,times,amsmath} 
\usepackage{url}
\usepackage{graphicx}
\usepackage{amsfonts}
\usepackage{wasysym}
\usepackage{latexsym}
\usepackage{color}
\usepackage{multirow}
\usepackage{amssymb}
\usepackage{epstopdf}

\usepackage{array}
\usepackage{multirow}
\usepackage{verbatim}
\usepackage{amstext}
\usepackage{amsmath}
\usepackage{rotating}
\usepackage{graphicx}
\usepackage{amssymb}
\usepackage{subfigure}
\usepackage[svgnames]{xcolor}
\usepackage{float}
\usepackage[ruled,vlined]{algorithm2e}
\usepackage{multirow}
\usepackage{url}
\usepackage{colortbl}
\usepackage{caption}
\usepackage{algorithmic}
\usepackage[ top=5cm, bottom=4cm, left=4cm , right=4cm]{geometry}

\begin{document}

\title { Bandit Models of Human Behavior: \\ Reward Processing in Mental Disorders}





\author{xx xx\inst{1}$^{,}$\inst{2} xx xx \inst{2}}

\authorrunning{Allesiardo et al.} 

\institute{xxx,\\
xxx\\
\and xxx}

\author{Djallel Bouneffouf, Irina Rish, Guillermo A. Cecchi}

\authorrunning{dj} 

\institute{IBM Thomas J. Watson Research Center\\
 Yorktown Heights, NY USA}

\maketitle 

\begin{abstract}
Drawing an inspiration from behavioral studies of human decision making, we propose here a general parametric framework for multi-armed bandit problem, which extends the standard Thompson Sampling approach to incorporate reward processing biases associated with several neurological and psychiatric conditions, including Parkinson's and Alzheimer's diseases, attention-deficit/hyperactivity disorder (ADHD), addiction, and chronic pain. We demonstrate empirically that the proposed parametric approach can often outperform the baseline Thompson Sampling on a variety of datasets. Moreover, from the behavioral modeling perspective, our parametric framework can be viewed as a first step towards a unifying computational model capturing reward processing abnormalities across multiple mental conditions.
\end{abstract}

\section{Introduction}
In daily-life decision making, from choosing a meal at a restaurant to deciding on a place to visit during a vacation, and so on, people often face the classical exploration versus exploitation dilemma, requiring them to choose between following a good action chosen previously (exploitation) and obtaining more information about the environment which can possibly lead to better actions in the future, but may also turn out to be a bad choice (exploration).

The exploration-exploitation trade-off is typically modeled as the {\em multi-armed bandit (MAB)} problem, stated as follows: given $N$ possible actions (``arms''), each associated with a fixed, unknown and independent reward probability distribution \cite {LR85,UCB}, an agent selects an action at each time point and receives a reward, drawn from the corresponding distribution, independently of the previous actions.

In order to better understand and model human decision-making behavior, scientists usually investigate reward processing mechanisms in healthy subjects \cite{perry2015reward}. However, neurogenerative and psychiatric disorders, often associated with reward processing disruptions, can provide an additional resource for deeper understanding of human decision making mechanisms. Furthermore, from the perspective of evolutionary psychiatry, various mental disorders, including
depression, anxiety, ADHD, addiction and even schizophrenia can be considered as ``extreme points'' in a continuous spectrum of behaviors and traits developed for various purposes during evolution, and  somewhat less extreme versions of those traits can be actually beneficial in specific environments (e.g., ADHD-like  fast-switching attention can be life-saving in certain environments, etc.). Thus, modeling decision-making biases and traits associated with various disorders may actually enrich the existing computational decision-making models, leading to potentially more flexible and better-performing algorithms.

Herein, we focus on reward-processing biases associated with several mental disorders, including Parkinson's and Alzheimer disease, ADHD, addiction and chronic pain. Our questions are: is it possible to extend standard stochastic bandit algorithms to mimic human behavior in such disorders? Can such generalized approaches outperform standard  bandit algorithms on specific tasks?

We show that both questions can be answered positively. We build upon the Thompson Sampling, a state-of-art approach to multi-arm bandit problem, and extend it to a parametric version which allows to incorporate various reward-processing biases known to be associated with particular disorders. For example, it was shown that (unmedicated) patients with  Parkinson's disease appear to learn better from negative rather than from positive rewards \cite{frank2004carrot};  another example is addictive behaviors which may be  associated with an inability  to forget strong  stimulus-response associations from the past, i.e. to properly discount past rewards \cite{redish2007reconciling}, and so on.
More specifically, \emph{we propose a  parameteric model which introduces  weights on incoming positive and negative rewards, and on reward  histories,  extending the standard parameter update rules in Bernoulli Thompson Sampling; tuning the parameter settings allows us to better capture  specific reward-processing biases.}

 Our empirical results demonstrate that the proposed approach outperforms the baseline Thompson Sampling on a variety of UCI benchmarks Furthermore, we show how parameter-tuning in the proposed model allows to mimic certain aspects of the behavior associated with mental disorders mentioned above, and thus may provide a valuable tool for improving our understanding of such disorders.



The rest of this paper is organized as follows. Section \ref{sec:related} reviews related work. Section \ref{sec:statement} describes the MAB model and the proposed algorithm. The experimental evaluation for different setting is presented in Section \ref{sec:experimental}. The last section concludes the paper and identifies directions for future works.

\vspace{-0.05in}
\section{Related Work}
\label{sec:related}
\vspace{-0.05in}
\subsection{Reward Processing in Mental Disorders}
\vspace{-0.05in}
The literature on the reward processing abnormalities in particular neurological and psychiatric disorders is quite extensive; below we summarize some of the recent developments in this fast-growing field.

\textbf{ Parkinson's disease (PD).} It is well-known that the neuromodulator dopamine plays a key role in reinforcement learning processes. PD patients, who have depleted dopamine in the basal ganglia, tend to have impaired performance on tasks that require learning from trial and error. For example, \cite{frank2004carrot} demonstrate that off-medication PD patients are better at learning to avoid choices that lead to negative outcomes than they are at learning
from positive outcomes, while dopamine medication typically used to treat PD symptoms reverses this bias.

\textbf{ Alzheimer's disease (AD).} This is the most common cause of dementia in the elderly and, besides memory impairment, it is associated with a variable degree of executive function
impairment and visuospatial impairment. As discussed in \cite{perry2015reward}, AD patients have decreased pursuit of rewarding behaviors, including loss of appetite; these changes are often secondary to apathy, associated with diminished reward system
activity. Furthermore, poor performance on certain tasks is correlated with memory impairments.

\textbf{ Frontotemporal dementia, behavioral variant (bvFTD). } Frontotemporal dementia (bvFTD) typically involves a progressive
change in personality and behavior including disinhibition, apathy, eating changes, repetitive or compulsive behaviors, and loss of empathy \cite{perry2015reward}, and it is hypothesized that those changes are associated with
abnormalities in reward processing. For example, changes in eating habits with a preference for sweet, carbohydrate rich foods and overeating in bvFTD patients can be associated with abnormally increased reward representation for food, or impairment in the negative (punishment) signal
associated with fullness.

\textbf{ Attention-deficit/hyperactivity disorder (ADHD).} Authors in \cite{luman2009does} suggest that the strength of the association between a stimulus and the corresponding response is more susceptible to degradation in ADHD patients, which suggests problems with storing the stimulus-response associations. Among other functions, storing the associations requires working memory capacity, which is often impaired in ADHD patients.

\textbf{ Addiction.} In \cite{redish2007reconciling}, it is demonstrated that patients suffering from addictive behavior are not able to forget the stimulus-response associations, which causes them to constantly seek the stimulus which generated such association.

\textbf{ Chronic pain.} In \cite{taylor2016mesolimbic}, it is suggested that chronic pain results in a hypodopaminergic (low dopamine) state that impairs motivated behavior, resulting into a reduced drive in chronic pain patients to pursue the rewards. Decreased reward response may underlie
a key system mediating the anhedonia and depression common in chronic pain.

A variety of computational  models was proposed for studying the disorders of reward processing in specific disorders, including, among others \cite{frank2004carrot,seeley2012frontotemporal,hauser2016computational,dezfouli2009neurocomputational,redish2007reconciling,hess2014beyond}.

However,   none of the above studies is proposing a unifying model that can represent a wide range of reward processing disorders; moreover, none of the above studies used the multi-arm bandit model simulating human online decision-making.

\vspace{-0.05in}
\subsection{Multi-Armed Bandit (MAB)}
\vspace{-0.05in}
The multi-armed bandit (MAB) problem models a sequential decision-making process, where at each time point a player selects an action from a given finite set of possible actions, attempting to maximize the cumulative reward over time.

MAB is frequently used in reinforcement learning to study the exploration/exploitation tradeoff, and is an active area of research since the 1950s. Optimal solutions have been provided using a stochastic formulation ~\cite {LR85,UCB}, or using an adversarial formulation ~\cite{AuerC98,AuerCFS02,BouneffoufF16}.
Recently, there has been a surge of interest in a Bayesian formulation \cite {chapelle2011empirical}, involving the algorithm known as Thompson sampling \cite {T33}. Theoretical analysis in \cite{AgrawalG12} shows that Thompson sampling for Bernoulli bandits asymptotically achieves the optimal performance limit.
Empirical analysis of Thompson sampling, including problems more complex than the Bernoulli bandit, demonstrates that its performance is highly competitive with other approaches \cite{chapelle2011empirical,DBouneffouf14}.

Psychological study done in \cite{schulz2015learning} shows that, instead of maximizing output by a deliberate mean-variance trade-off, participants approach dynamic decision-making problems by utilizing a probability matching heuristic. Thus, their behavior is better described by the Thompson sampling choice rule than by the Upper Confidence Bound (UCB) approach \cite{UCB}.
However, none of the above studies bandit models of the behavior of patients with mental disorders and impaired reward processing.

To the best of our knowledge, this work is the first one to propose a generalized version of Thompson Sampling algorithm which incorporates a range of reward processing biases associated with various mental disorders and shows how different parameter settings of the proposed model lead to behavior mimicking a wide range of impairments in multiple neurological and psychiatric disorders. Most importantly, our bandit algorithm based on generalization of Thompson sampling outperforms the baseline method on multiple datsasets.

\vspace{-0.05in}
\section{Background and Definitions }
\vspace{-0.05in}
\label{sec:statement}

\noindent{\bf The Stochastic Multi-Armed Bandit.} Given a slot machine with $N$ arms representing potential actions, the player must chose one of the arms to play at each time step $t= 1,2,3,...,T$. Choosing an arm $i$ yields a random real-valued reward according to some fixed (unknown) distribution with support in $[0,1]$.
The reward is observed immediately after playing the arm. The MAB algorithm must decide which arm to play at each time step $t$, based on the outcomes during the previous $t-1$ steps.

Let $\mu_i$ denote the (unknown) expected reward for arm $i$. The goal is to maximize the expected total reward during $T$ iterations, i.e., $E[\sum^T_{t=1} \mu_{i(t)}]$, where $i(t)$ is the arm played in step $t$, and the expectation is over the random choices of $i(t)$ made by the algorithm. We could also use the equivalent performance measure known as the expected total regret, i.e. the amount of total reward lost because of playing according to a specific algorithm rather than choosing the optimal arm in each step.

The expected total regret is formally defined as:
 \vspace{-0.01in}
\begin{eqnarray}
E[R(T)] =E[\sum^T_{t=1}(\mu^*-\mu_{i_{(t)}})]= \sum_i \Delta_i E[k_i(T)].
\end{eqnarray}
 \vspace{-0.01in}
where $\mu^*:= max_i \mu_i$,  $ \Delta_i:= \mu^*-\mu_i$, and $k_i(t)$ denote the number of times arm $i$ has been played up to step $t$.


\noindent{\bf Thompson Sampling.}
Thompson sampling (TS) \cite{thompson1933likelihood}, also known as
Basyesian posterior sampling, is a classical approach to multi-arm bandit problem, where the reward $r_{i}(t)$ for choosing an arm $i$ at time $t$ is assumed to follow a distribution $Pr(r_{t}|\tilde{\mu})$ with the parameter $\tilde{\mu}$. Given a prior $Pr(\tilde{\mu})$ on these parameters, their posterior distribution is given by the Bayes rule, $Pr(\tilde{\mu}|r_{t}) \propto Pr(r_{t}|\tilde{\mu}) Pr(\tilde{\mu})$ \cite{AgrawalG12}.

A particular case of the Thompson Sampling approach, presented in Algorithm 1, assumes a Bernoulli bandit problem, with rewards being 0 or 1, and the parameters following the Beta prior.
TS initially assumes arm $i$ to have prior $Beta(1, 1)$ on $\mu_i$ (the probability of success). At time $t$, having observed $S_i(t)$ successes (reward = 1) and $F_i(t)$ failures (reward = 0), the algorithm updates the distribution on $\mu_i$ as $Beta(S_i(t), F_i(t))$. The algorithm then generates independent samples $\theta_i(t)$ from these posterior distributions of the $\mu_i$, and selects the arm with the largest sample value.

\vspace{-0.1in}
\begin{algorithm}[th]
 \caption{Thompson Sampling }
\label{alg:TS}
\begin{algorithmic}[1]
 \STATE {\bfseries }\textbf{Foreach} arm $i= 1, . . . , K$
 \STATE {\bfseries } \quad set
 $ S_i(t) = 1$, $F_i(t) = 1$
 \STATE {\bfseries }\textbf{End for}
 \STATE {\bfseries }\textbf{Foreach} $t = 1, 2, . . . ,T$ \textbf{do}
 \STATE {\bfseries } \quad \textbf{Foreach} $i = 1, 2, . . . ,K$ \textbf{do}
 \STATE {\bfseries } \quad \quad Sample $\theta_i(t)$ from $Beta(S_i(t), F_i(t))$ 
 \STATE {\bfseries } \quad \textbf{End do}
 \STATE {\bfseries }\quad Play arm $i_t= argmax_i \theta_i(t)$, obtain reward $r(t)$
 \STATE {\bfseries } \quad \textbf{if} $r(t)=1$, \textbf{then}
 \STATE {\bfseries } \quad \quad $S_i(t)=S_i(t)+1$
 \STATE {\bfseries } \quad \textbf{else} $F_i(t)=F_i(t)+1$
 \STATE {\bfseries }\textbf{End do}
 \end{algorithmic}
\end{algorithm}
  \vspace{-0.05in}

\section{Proposed Approach: Human-Based Thompson Sampling }
\vspace{-0.05in}
\label{subsec:UCB}
We will now introduce a more general rule for updating the parameters of Beta distribution in steps 10 and 11 of the Algorithm 1; this parameteric rule incorporates weights on the prior and the current number of successes and failures, which will allow to model a wide range of reward processing biases associated with various disorders.
More specifically, the proposed Human-Based Thompson Sampling (HBTS), outlined in Algorithm \ref{alg:HBTS}, replaces binary incremental updates in lines 10 and 11 of TS (Algorithm 1) with their corresponding weighted version (lines 10 and 11 in Algorithm \ref{alg:HBTS}), using the four weight parameters:
 $\tau $ and $\phi$ are the weights of the previously accumulated positive and negative rewards, respectively, while $\alpha$ and $\beta$ represent the weights on the positive and negative rewards at the current iteration.

\vspace{-0.1in}
\begin{algorithm}[th]
 \caption{Human-Based Thompson Sampling (HBTS)}
\label{alg:HBTS}
\begin{algorithmic}[1]
 \STATE {\bfseries }\textbf{Foreach} arm $i= 1, . . . , K$
 \STATE {\bfseries } \quad set
 $ S_i(t) = 1$, $F_i(t) = 1$
 \STATE {\bfseries }\textbf{End for}
 \STATE {\bfseries }\textbf{Foreach} $t = 1, 2, . . . ,T$ \textbf{do}
 \STATE {\bfseries } \quad \textbf{Foreach} $i = 1, 2, . . . ,K$ \textbf{do}
 \STATE {\bfseries } \quad \quad Sample $\theta_i(t)$ from $Beta(S_i(t), F_i(t))$ 
 \STATE {\bfseries } \quad \textbf{End do}
 \STATE {\bfseries }\quad Play arm $i_t= argmax_i \theta_i(t)$, obtain reward $r(t)$
 \STATE {\bfseries } \quad \textbf{if} $r_i(t)=1$, \textbf{then}
 \STATE {\bfseries } \quad \quad $S_i(t)=\tau S_i(t)+ \alpha r_i(t) $ \STATE {\bfseries } \quad \quad \textbf{else} $F_i(t)=\phi F_i(t)+ \beta (1-r_i(t)) $
 \STATE {\bfseries }\textbf{End do}
 \end{algorithmic}
\end{algorithm}
  \vspace{-0.1in}

\subsection{Reward Processing Models with Different Biases}
\vspace{-0.05in}
In this section we describe how specific constraints on the model parameters in the proposed algorithm can yield different reward processing biases discussed earlier, and introduce several  instances of the HBTS model, with parameter settings reflecting particular biases.   The parameter settings are summarized in Table 2, where we use list our models associated with  specific disorders.

It is important to underscore  that the above models should be viewed as only  a first step towards a unifying approach to reward processing disruptions, which requires further extensions, as well as tuning and validation on human subjects. Our main goal is to  demonstrate the promise of our parametric approach  at capturing certain decision-making biases, as well as its computational advantages over the standard TS, due to the increased generality and flexibility facilitated by multi-parametric formulation.
Note that the standard Thompson sampling (TS) approach correspond to setting the four (hyper)parameters used in our model to 1. Next, we introduce the model which incorporates some mild forgetting of the past rewards or losses, using 0.5 weights, just as an example, and calibrating the other models with respect to this one; we refer to this model as M for ``moderate'' forgetting,
which serves here as a proxy for somewhat ``normal'' reward processing, without extreme reward-processing biases associated with disorders.
We will use the subscript $M$ to denote the parameters of this model.

\begin{table}[t]
\centering
\caption{Algorithms parameters}
\resizebox{0.62\columnwidth}{!}{
 \begin{tabular}{ l | c | r | c | r }
 UCI Datasets & $\tau$   & $\alpha$     & $\phi$      & $\beta$ \\ \hline
  AD (addiction)   & $1 \pm 0.1$   & $1 \pm 0.1$    & $0.5 \pm 0.1$   & $1 \pm 0.1$ \\
  ADHD  & $0.2\pm 0.1$  & $1 \pm 0.1$    & $0.2 \pm 0.1$   & $1 \pm 0.1$ \\
  AZ (Altzheimer's)   & $0.1 \pm 0.1$  & $1 \pm 0.1$    & $0.1 \pm 0.1$   & $1 \pm 0.1$ \\
  CP (chronic pain)   & $0.5 \pm 0.1$  & $0.5 \pm 0.1$    & $1 \pm 0.1$    & $1 \pm 0.1$ \\
  bvFTD   & $0.5 \pm 0.1$  & $100 \pm 10$    & $0.5 \pm 0.1$   & $1 \pm 0.1$ \\
  PD (Parkinson's)   & $0.5 \pm 0.1$  & $1 \pm 0.1$    & $0.5\pm 0.1$   & $100\pm 10$\\
 M (``moderate'')   & $0.5 \pm 0.1$  & $1 \pm 0.1$    & $0.5 \pm 0.1$   & $1 \pm 0.1$ \\
  TS   & 1        & 1         & 1         & 1\\
 \end{tabular}
 }
  \vspace{-0.2in}
 \label{table:Synthetic}
 \end{table}

We will now introduced several models inspired by certain reward-processing biases in a range of mental disorders. {\em It is important to note that, despite using disorder names for these models, we are not claiming  that they provide accurate models of the corresponding disorders, but rather disorder-inspired versions of our general parametric family of models.}

\textbf{ Parkinson's disease (PD).} Recall that PD patients are typically better at learning to avoid negative outcomes than at learning to achieve positive outcomes \cite{frank2004carrot}; one way to model this is to over-emphasize negative rewards, by placing a high weight on them, as compared to the reward processing in healthy individuals. Specifically, we will assume the
parameter $\beta$ for PD patients to be much higher than normal $\beta_M$ (e.g., we use $\beta=100$ here), while the rest of the parameters will be in the same range for both healthy and PD individuals.

\textbf{Frontotemporal Dementia (bvFTD).}
Patients with bvFTD are prone to overeating which may represent increased reward representation. To model this impairment in bvFTD patients, the parameter of the model could be modified as follow:
$\alpha_M <<\alpha$ (e.g.,  $\alpha=100$ as shown in Table 2), where $\alpha$ is the parameter of  the bvFTD model has, and the rest of these parameters are equal to the normal one.

\textbf{Alzheimer's disease (AD).} To model apathy in  patients with Altzheimer's, including downplaying rewards and losses, we will assume that the parameters $\phi$ and $\tau $ are  somewhat smaller than normal, $ \phi < \phi_M$ and $\tau < \tau_M $ (e.g,  set to 0.1 in Table 2), which models the tendency to  forget both positive and negative rewards.

\textbf{ADHD.} Recall that ADHD may be involve impairments in storing stimulus-response associations. In our ADHD model, the parameters $\phi$ and $\tau $ are smaller than normal, $\phi_M > \phi$ and $\tau_M > \tau$, which models forgetting of both positive and negative rewards. Note that while this model appears similar to Altzheimer's model described above, the forgetting factor will be  less pronounced, i.e. the $\phi$ and $\tau $ parameters are larger than those of the  Altzheimer's model (e.g., 0.2 instead of 0.1, as shown in Table 2).

\textbf{Addiction.} As mentioned earlier, addiction is associated with inability to properly forget (positive) stimulus-response associations; we model this by setting the weight on previously accumulated positive reward (``memory'' )  higher than normal, $\tau >\tau_M $, e.g. $\tau = 1$, while $\tau_M = 0.5$.

\textbf{Chronic Pain.} We model the reduced responsiveness to rewards in chronic pain by setting $ \alpha < \alpha_M $ so there is a decrease in the reward representation, and $\phi > \phi_M $ so the negative rewards are not forgotten (see table 2).

Of course, the above models should be treated only as first approximations of the reward processing  biases in mental disorders, since the actual changes in reward processing are much more complicated, and the parameteric setting must be learned from actual patient data, which is a nontrivial direction for future work. Herein, we simply consider those models as specific variations of our general method, inspired by certain aspects of the corresponding diseases, and focus primarily on the computational aspects of our algorithm, demonstrating that the proposed parametric extension of TS can learn better than the baseline TS due to added flexibility.

\vspace{-0.05in}
\section{Empirical Evaluation}
\vspace{-0.05in}
\label{sec:experimental}
In order to evaluate the proposed framework empirically and compare its performance with the standard Thompson Sampling, we used the following four classification datasets from the UCI Machine Learning Repository\footnote{https://archive.ics.uci.edu/ml/datasets.html}:
Covertype, CNAE-9, Internet Advertisements and Poker Hand. A brief summary of the datasets is listed in Table \ref{table:Synthetic1}.

\begin{table}[t]
\centering
\caption{Datasets}
\resizebox{0.45\columnwidth}{!}{
 \begin{tabular}{ l | c | r }
 UCI Datasets  & Instances  & Classes \\ \hline
  Covertype  & 581 012  & 7\\
  CNAE-9   & 1080  & 9\\
  Internet Advertisements & 3279  & 2\\
  Poker Hand  & 1 025 010  & 9\\
 \end{tabular}
 }
   \vspace{-0.2in}
 \label{table:Synthetic1}
 \end{table}

In order to simulate an infinite  data stream, we draw samples randomly without replacement, from each dataset,   restarting the process each time we draw the last sample. In each round, the algorithm
receives the reward 1 if the instance is classified correctly, and 0 otherwise. We compute the total number of classification errors as a performance metric. Note that we do not  use the features (context) here, as we try to simulate the classical multi-arm bandit environment (rather than contextual bandit), and use the   class labels only.  As the result, {\em we obtain  a non-stationary environment}, since even if $P(reward|context)$ is fixed,  switching from a sample to a sample (i.e., from a context to a context) results into different  $P(reward)$ at each time point.

\begin{table*}[ht!]
\centering
\caption{Average	Results}
\resizebox{0.99\columnwidth}{!}{
 \begin{tabular}{ l | l | l | l | l | l | l | l | l}
 	            & Addiction    & ADHD     & Alzheimer's  & Chronic Pain & bvFTD  & Parkinson & M    & TS \\ \hline
 Datasets \\ \hline
Positive Environment    & $\textbf{51.46}$     & $ 52.35 $  & $52.53$    & $52.88$ & $59.16$ & $56.23$  & $52.64$ & $55.62$ \\
Negative Environment    & $62.83$     & $ 55.06 $  & $55.54$    & $55.48$ & $56.03$ & $56.21$  & $\textbf{52.74}$ & $61.21$ \\
Normal Reward Environment & $52.81$     & $ 53.68 $  & $51.48$    & $53.11$ & $\textbf{49.55}$ & $58.01$  & $50.92$ & $56.95$ \\
 \hline
 \end{tabular}
 }
  \vspace{-0.15in}
 \label{table:AccuSyn}
 \end{table*}

\begin{table*}[ht!]
\centering
\caption{Positive-reward Environment}
\resizebox{0.99\columnwidth}{!}{
 \begin{tabular}{ l | l | l | l | l | l | l | l | l}
 	           & Addiction    & ADHD   & Alzheimer's  & Chronic Pain  & bvFTD & Parkinson & M & TS \\ \hline
 Datasets \\ \hline
 Internet Advertisements & $34.06\pm0.34$  & $\textbf{31.85}\pm\textbf{3.51}$	 & $32.40\pm 1.91$	& $32.96\pm\ 1.66$ & $55.67\pm1.68$ & $43.61\pm1.51$ & $37.69\pm1.88$  & $38.34\pm1.77$\\
 CNAE-9 	       & $40.25 \pm 0.85$ & $\textbf{39.89}\pm \textbf{2.70}$ & $40.08\pm3.69$	& $39.94\pm 0.73$ & $40.14\pm2.33$ & $40.28\pm2.27$ & $40.16\pm1.99$ & $40.06\pm1.66$ \\
Covertype        & $65.04\pm 0.52$ & $66.5\pm 0.75$  & $66.75\pm1.52$ & $69.49\pm1.75$ & $70.62\pm1.73$ & $68.05\pm1.72$ & $\textbf{65.01}\pm \textbf{1.75}$ & $67.08\pm1.23$\\
Poker Hand 	       & $\textbf{66.5}\pm \textbf{0.24}$ & $ 71.18 \pm 0.12$ & $70.19\pm1.87$ & $69.14\pm2.57$ & $70.26\pm0.81$ & $73\pm1.87$ & $67.73\pm1.87$ & $77.03\pm1.87$ \\
 \hline
 \end{tabular}
 }
   \vspace{-0.15in}
 \label{table:AccuSyn_pos}
 \end{table*}

 \begin{table*}[ht!]
\centering
\caption{Negative-reward Environment}
\resizebox{0.99\columnwidth}{!}{
 \begin{tabular}{ l | l | l | l | l | l | l | l | l}
 	           & Addiction    & ADHD   & Alzheimer's  & Chronic Pain & bvFTD & Parkinson & M & TS \\ \hline
 Datasets \\ \hline
 Internet Advertisements & $41.346\pm0.21$  & $37.833 \pm 1.20$ & $40.76\pm 1.93$	& $ 41.08\pm\ 1.64$ & $42.633\pm1.23$ & $41.4\pm1.17$ & $ \textbf{33.22}\pm \textbf{1.7}$  & $38.19\pm1.6$\\
 CNAE-9  & $40.248 \pm 0.35$ & $39.97\pm0.20$ & $39.89\pm3.49$	& $40.27\pm 0.23$ & $39.89\pm1.33$ & $39.95\pm1.73$ & $\textbf{39.96}\pm \textbf{1.33}$ & $40.02\pm1.11$ \\
Covertype        & $73.26 \pm 0.30$ & $ 71.28 \pm 0.32$  & $71.35\pm 1.75$ & $71.45\pm1.87$ & $71.34\pm1.87$ & $\textbf{70.5}\pm\textbf{1.8}$ & $70.05\pm1.87$ & $69.93 \pm0.83$\\
Poker Hand 	       & $96.51 \pm 0.35$ & $ 71.18 \pm 0.22$ & $70.19 \pm2.77$ & $69.14\pm0.88$ & $70.26\pm1.19$ & $73\pm1.87$ & $\textbf{67.73}\pm \textbf{1.51}$ & $96.71\pm1.16$ \\
 \hline
 \end{tabular}
 }
   \vspace{-0.15in}
 \label{table:AccuSyn_neg}
 \end{table*}

\begin{table*}[ht!]
\centering
\caption{Normal Reward Environment}
\resizebox{0.99\columnwidth}{!}{
 \begin{tabular}{ l | l | l | l | l | l | l | l | l}
 	           & Addiction    & ADHD   & Alzheimer's  & Chronic pain  & bvFTD & Parkinson & M & TS \\ \hline
 Datasets \\ \hline
 Internet Advertisements & $32.28\pm 0.20$  & $36.8 \pm 1.28$	 & $36.56\pm 1.63$	& $35.53\pm\ 1.43$ & $\textbf{28.59}\pm1.76$ & $44.69\pm1.85$ & $33.65\pm1.81$ & $37.71\pm 0.66$\\
 CNAE-9 	       & $40.16 \pm 0.38$ & $40.09\pm0.31$ & $39.99\pm3.01$	& $\textbf{39.75}\pm \textbf{0.28}$ & $40.13\pm1.81$ & $40.25\pm1.71$ & $39.86\pm1.10$ & $39.78\pm0.80$ \\
Covertype        & $73.54\pm 0.31$ & $ 64.27\pm 0.30$  & $\textbf{63.54}\pm\textbf{1.30}$ & $68.69\pm1.84$ & $63.69\pm1.85$ & $72.67\pm1.82$ & $64.61\pm0.8$ & $64.63\pm1.87$\\
Poker Hand 	       & $\textbf{65.29}\pm 0.33$ & $ 73.57 \pm 0.33$  & $65.83\pm2.68$ & $68.49\pm0.92$ & $65.69\pm1.01$ & $74.44\pm1.07$ & $65.58\pm1.62$ & $85.71\pm1.09$ \\
 \hline
 \end{tabular}
 }
   \vspace{-0.15in}
 \label{table:AccuSyn_norm}
 \end{table*}

In order to test the ability of our  models to reflect  decision-making biases in various disorders, as well as to  evaluate the advantages of our model in comparison with the baseline TS, under different test conditions, we consider the following   settings:

- Positive reward environment: we modify the reward function so that the agent receives only positive rewards (the lines 11 is not executed). This environment allows us to evaluate how our models deal with positive reward.

- Negative reward environment: we modify the reward function so that the agent receives only negative rewards (the lines 10 is not executed). This environment helps to evaluate the negative-reward processing  by our models.

- Normal environment: the agent can see both negative and positive rewards.

The average error rate results on the UCI datasets, for each type of the environment, and over 10 runs of each algorithm, are shown in Table \ref{table:AccuSyn}. We compute the error rate by dividing the total accumulated regret by the number of iterations. The best results for each dataset are shown in bold. Note that our parametric approach always outperforms the standatrd TS method: AD (addiction) model is best in positive reward environment, M (moderate) version is best in negative environment, and
bvFTD happens to outperform other models in regular (positive and negative) reward environment. While further  modeling and validation on human subjects may be required to validate neuroscientific value of the proposed models, they clearly demonstrate computational advantages over the classical TS approach for the bandit problem.

We now present the detailed results for all algorithms and for each of the three environments, in Tables \ref{table:AccuSyn_pos}, \ref{table:AccuSyn_neg}, and \ref{table:AccuSyn_norm}. Lowest errors for each dataset (across each row) are again shown in bold. Note that, {\em in all three environments, and for each of the four datasets, the baseline Thompson Sampling was always inferior to the proposed parametric family of methods}, for each specific settings, different  versions of our HBTS framework were performing best.

 \vspace{-0.1in}
\subsubsection{Positive Reward Environment.}
\vspace{-0.05in}
Table \ref{table:AccuSyn_pos} summarizes the results for positive reward setting. Note that most versions of the proposed approach frequently outperform  the standard Thompson sampling. ADHD model yields best results on two datasets out of four, while AD (addiction) and M (moderate) models are best at one of each remaining datasets, respectively.

Note that PD (Parkinson's) and bvFTD (behavioral-variant fronto-teporal dementia) yield the worst results on most datasets. The behavior of PD model is therefore consistent with the literature on Parkinson's disease, which suggests, as mentioned earlier, that Parkinson's patients do not learn as well from positive rewards as they do from negative ones.

Ranking the algorithms with respect to their mean error rate, we note that the top three performing algorithms were AD (addiction), ADHD and AZ (Alzheimer's), in that order. One can hypothesise that these observations are consistent with the fact that those disorders did not demonstrate such clear impairment in learning from positive rewards as, for example, Parkinson's.

 \vspace{-0.1in}
\subsubsection{Negative Reward Environment.}
\vspace{-0.05in}
As shown in Table \ref{table:AccuSyn_neg}, for negative reward environment, we again observe that  the proposed algorithms alwyas work better than the state of the art Thompson sampling.

Overall, M (moderate) model performs best in this environment, on three out of four datasets. Note that PD (Parkinson's) and CP (chronic pain) models outperform many other models, performing much better with negative rewards than they did woith the positive ones, which is consistent with the literature discussed before. AD (addiction) is the worst-performing out of HBTS algorithms, which may relate to its bias towards  positive-reward driving learning, but impaired ability to learn from negative rewards.

Ranking the algorithms with respect to their mean error rate, we note that the two best-performing algorithms were ADHD and AZ (Alzheimer's), in that order.

 \vspace{-0.1in}
\subsubsection{Normal Reward Environment.}
\vspace{-0.05in}
Similarly to the other two environments, the baseline Thompson Sampling is always inferior to the proposed algorithms, as shown in Table \ref{table:AccuSyn_norm}).
Interestingly,  model M was never a winner, either, and different disorder models performed best for different data sets. PD and CP show worst performance, suggesting that negative-reward driven learning is impairing.

 \vspace{-0.05in}
\section{Conclusions}
 \vspace{-0.05in}
\label{sec:Conclusion}
This paper proposes a novel parametric family of algorithms for multi-arm bandit problem, extending the classical Thompson Sampling approach to model
a wide range of potential reward processing biases. Our approach  draws an inspiration from extensive literature on decision-making behavior in neurological and psychiatric disorders stemming from disturbances of the reward processing system.
The proposed model is shown to consistently outperform the baseline Thompson Sampling method, on all data and experiment settings we explored,
demonstrating better adaptation to each domain due to high flexibility of our multi-parameter model which allows to tune the weights on incoming positive and negative rewards, as well as the weights on memories about the prior reward history.
Our empirical results support multiple prior observations about reward processing biases in a range of mental disorders, thus indicating the potential of the proposed model and its future extensions to capture reward-processing aspects across various neurological and psychiatric conditions. Our future work directions include  extending our model to the more realistic contextual bandit setting, as well as testing the model  on human decision making data.


\vspace{-2mm}
\bibliographystyle{splncs}
\small
\bibliography{biblio}

\end{document}